\documentclass{article} %
\usepackage{iclr2022_conference,times}

\usepackage{amsmath,amsfonts,bm}

\def\eqref#1{equation~\ref{#1}}

\def\1{\bm{1}}

\def\rc{{\textnormal{c}}}

\DeclareMathAlphabet{\mathsfit}{\encodingdefault}{\sfdefault}{m}{sl}
\SetMathAlphabet{\mathsfit}{bold}{\encodingdefault}{\sfdefault}{bx}{n}

\usepackage[T1]{fontenc}    %
\usepackage[colorlinks=true,citecolor=blue]{hyperref}
\usepackage{booktabs}       %
\usepackage{amsfonts}       %
\usepackage{nicefrac}       %
\usepackage{microtype}      %
\usepackage{xcolor}  
\usepackage{color, colortbl}
\usepackage{fancyvrb}
\usepackage{epsfig}
\usepackage{graphicx}
\usepackage{wrapfig}
\usepackage{adjustbox}
\usepackage{mathtools}
\usepackage{caption}
\usepackage{xspace}
\usepackage{algorithm}
\usepackage{listings}
\usepackage{import}
\usepackage{subcaption}

\definecolor{Gray}{gray}{0.9}

\def\super{SIFAR\xspace}

\def\superS{SIFAR-B-7\xspace}

\def\superM{SIFAR-B-12\xspace}

\def\superL{SIFAR-B-14\xspace}

\def\superMM{SIFAR-B-12$\dagger$\xspace}

\def\superLL{SIFAR-B-14$\dagger$\xspace}

\def\superHH{SIFAR-B-12$\ddagger$\xspace}

\def\LsuperLL{SIFAR-L-14$\dagger$\xspace}

\def\LsuperHH{SIFAR-L-12$\ddagger$\xspace}

\def\ours{SIFAR\xspace}
\def\Swin{Swin Transformer\xspace}

\def\zapcolorreset{\let\reset@color\relax\ignorespaces}
\def\colorrows#1{\noalign{\aftergroup\zapcolorreset#1}\ignorespaces}

\newcommand{\mycomment}[1]{}

\title{
Can an Image Classifier Suffice for Action Recognition?
}

\author{Quanfu Fan\thanks{Equal contribution.}, \ Chun-Fu (Richard) Chen\footnotemark[1], \ Rameswar Panda\footnotemark[1]\\
MIT-IBM Watson AI Lab \\
\texttt{qfan@us.ibm.com, chenrich@us.ibm.com, rpanda@ibm.com}
}

\iclrfinalcopy
\begin{document}

\maketitle

\begin{abstract}
We explore a new perspective on video understanding by casting the video recognition problem as an image recognition task. Our approach rearranges input video frames into super images, which allow for training an image classifier directly to fulfill the task of action recognition, in exactly the same way as image classification. With such a simple idea, we show that transformer-based image classifiers alone can suffice for action recognition. In particular, our approach demonstrates strong and promising performance against SOTA methods on several public datasets including Kinetics400, Moments In Time, Something-Something V2 (SSV2), Jester and Diving48. We also experiment with the prevalent ResNet image classifiers in computer vision to further validate our idea. The results on both Kinetics400 and SSV2 are comparable to some of the best-performed CNN approaches based on spatio-temporal modeling. Our source codes and models are available at \url{https://github.com/IBM/sifar-pytorch}.
\end{abstract}

\section{Introduction}
\label{sec:intr}

\begin{wrapfigure} {r}{0.4\textwidth}
    \centering
    \includegraphics[width=1.\linewidth]{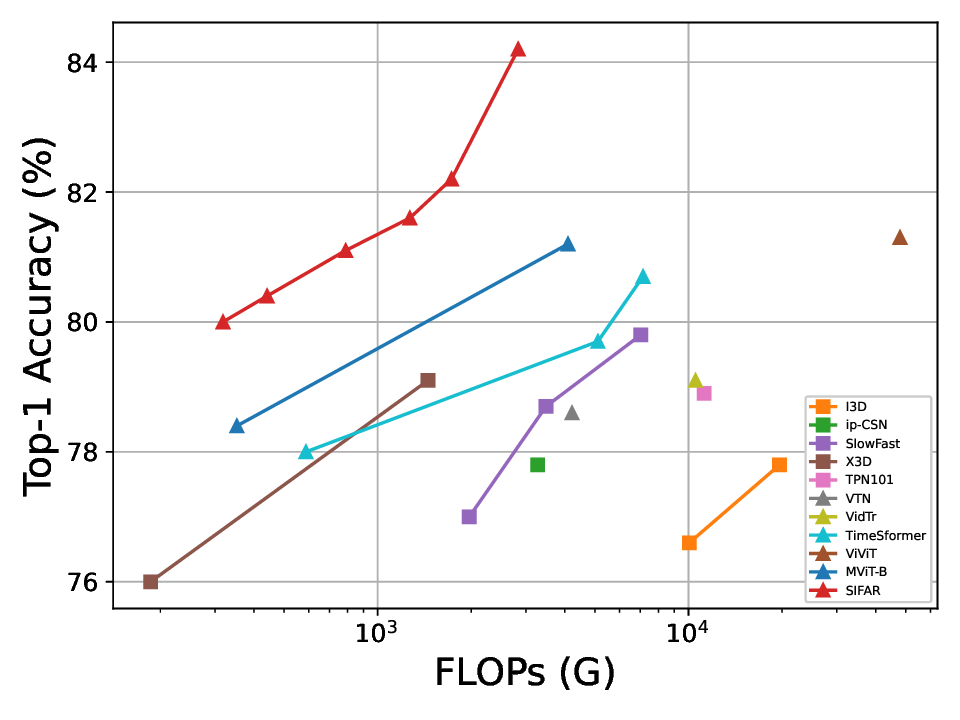}
    \caption{\small{Comparison of our proposed SIFAR (red) with SOTA approaches for action recognition on Kinetics400 dataset.}}
    \label{fig:k400_results} 
     \vspace{-3mm}
\end{wrapfigure}
The recent advances in convolutional neural networks (CNNs)~\citep{ResNet_He_2016_CVPR, efficientnet_pmlr_tan_19}, along with the availability of large-scale video benchmark datasets~\citep{Kinetics:kay2017kinetics, Moments:monfort2019moments, DBLP:journals/corr/abs-2006-13256},  have significantly improved action recognition, one of the fundamental problems of video understanding. 
Many existing approaches for action recognition naturally extend or borrow ideas from image recognition. At the core of these approaches is \textit{spatio-temporal modeling}, which regards time as an additional dimension and jointly models it with space by extending image models (i.e., 3D CNNs)  ~\citep{C3D:Tran2015learning, I3D:carreira2017quo, X3D_Feichtenhofer_2020_CVPR} or fuses temporal information with spatial information processed separately by 2D CNN models \citep{TSM:lin2018temporal, bLVNetTAM}. CNN-based approaches demonstrate strong capabilities in learning saptio-temporal feature representations from video data. %

Videos present long-range pixel interactions in both space and time. It's known in approaches like non-local networks~\citep{Nonlocal:Wang2018NonLocal} that modeling such relationships helps action recognition. The recently emerging Vision Transformers (ViTs) naturally own the strength of capturing long-range dependencies in data, making them very suitable for video understanding. Several approaches~\citep{TimeSformer_Bertasius_2021vc, VidTr,arnab2021vivit} have applied ViTs for action recognition and shown better performance than their CNN counterparts. However, these approaches are still following the conventional paradigm of video action recognition, and perform temporal modeling in a similar way to the CNN-based approaches using dedicated self-attention modules.

In this work, we explore a different perspective for action recognition by casting the problem as an image recognition task. We ask if it is possible to model temporal information with ViT directly without using dedicated temporal modules. In other words, can an image classifier alone suffice for action recognition? 
To this end, we first propose a simple idea to turn a 3D video into a 2D image. Given a sequence of input video frames, we rearrange them into a super image according to a pre-defined spatial layout, as illustrated in Fig.~\ref{fig:overview}.  The super image encodes 3D spatio-temporal patterns in a video into 2D spatial image patterns. We then train an image classifier to fulfill the task of action recognition, in exactly the same way as image classification. Without surprise, based on the concept of super images, any image classifier can be re-purposed for action recognition. For convenience, we dub our approach \textit{SIFAR}, short for \textbf{S}uper \textbf{I}mage \textbf{f}or \textbf{A}ction \textbf{R}ecognition. 

 We validate our proposed idea by using \Swin~\citep{Swin_Liu_2021tq}, a recently developed vision transformer that has demonstrated good performance on both image classification and object detection. Since a super image has a larger size than an input frame,  we modify the \Swin to allow for full self-attention in the last layer of the model, which further strengthens the model's ability in capturing long-range temporal relations across frames in the super image. With such a change, we show that \super produces strong performance against the existing SOTA approaches (Fig.~\ref{fig:k400_results}) on several benchmark datsets including Kinetics400~\citep{Kinetics:kay2017kinetics}, Moments in Time~\citep{Moments:monfort2019moments}, Something-Something V2 (SSV2)~\cite{Something:goyal2017something}, Jester~\citep{Materzynska_2019_ICCV} and Diving48~\citep{Li_2018_ECCV}. Our proposed \super also enjoys efficiency in computation as well as in parameters. We further study the potential of CNN-based classifiers directly used for action recognition under the proposed \ours framework. Surprisingly, they achieve very competitive results on Kinetics400 dataset against existing CNN-based approaches that rely on much more sophisticated spatio-temporal modeling. 
 Since $3\times3$ convolutions focus on local pixels only, CNN-based \super handles temporal actions on Something-Something less effectively. We experiment with larger kernel sizes to expand the temporal receptive field of CNNs, which substantially improves the CNN-based \super by $4\%-6.8\%$ with ResNet50.

\ours brings several advantages compared to the traditional spatio-temporal action modeling. Firstly, it is simple but effective. With one single line of code change in PyTorch, SIFAR can use any image classifier for action recognition. We expect that similar ideas can also work well with other video tasks such as video object segmentation~\citep{DBLP:journals/corr/abs-2101-08833}. Secondly, \super makes action modeling easier and more computationally efficient as it doesn't require dedicated modules for temporal modeling. Nevertheless, we do not tend to underestimate the significance of temporal modeling for action recognition. Quite opposite, \super highly relies on the ability of its backbone network to model long-range temporal dependencies in super images for more efficacy. Lastly, but not the least, the perspective of treating action recognition the same as image recognition unleashes many possibilities of reusing existing techniques in a more mature image field to improve video understanding from various aspects. For example, better model architectures~\citep{efficientnet_pmlr_tan_19}, model pruning~\citep{liu2017learning} and interpretability~\citep{desai2020ablationcam}, to name a few.

\mycomment{
\QF{
To-do-list
\begin{itemize}
    \item Improving st2st results (bottleneck, dilated convolution)?
    \item CNN-based results (try models with dilation, long-range dependencies?, do R50 and R101, check the performance gap, better to find imagenet22k models)*
    
    \item use cases to showcase advantages of the proposed approach (one or two examples)*
    
    \rc{a: try image augmentation?}
    another example ?
    
    \item pretrain on other models (i.e. tiny and large ones?) try to acheive SOTA results on kinetics400 and moments
    
    \item effects of position-embedding (do this on frame basis ) 
    \item  visualization
    
    \item SWING transformer with separate temporal modeling
    \item pretrain on imagenet1K models
    \item TWIN or our own approach ?

\end{itemize}
}}
\section{Related Work}
\label{sec:literature}

\textbf{Action Recognition from a Single Image.} One direction for video action recognition is purely based on a single image~\citep{davis1997mei, Zhao_2017_ICCV, 8658386, bilen2016dynamic}. In ~\citep{davis1997mei}, multiple small objects are first identified in a still image and then the target action is inferred from the relationship among the objects. 
Other approaches such as~\citep{8658386} propose to predict the missing temporal information in still images and then combine it with spatial information for action classification.
There are also approaches that attempt to summarize RGB or motion information in a video into a representative image for action recognition, e.g., motion-energy image (MEI)~\citep{davis1997mei}, Dynamic Image Network~\citep{bilen2016dynamic},  Informative Frame Synthesis (IFS)~\citep{Qiu_2021_ICCV}, Adaptive Weighted Spatio-temporal Distillation (AWSD)~\citep{Tavakolian_2019_ICCV} and Adversarial Video Distillation (AVD)~\citep{DBLP:journals/corr/abs-1907-05640}. 
Nonetheless, our approach does not attempt to understand a video from a single input image or a summarized image. Instead our proposed approach composites the video into a super image, and then classifies the image with an image classifier directly.

\textbf{Action Recognition with CNNs.} Action recognition is dominated by CNN-based models recently~\citep{SlowFast:feichtenhofer2018slowfast, I3D:carreira2017quo, bLVNetTAM, X3D_Feichtenhofer_2020_CVPR, chen2021deep, TSM:lin2018temporal, TSN:wang2016temporal, TRN:zhou2018temporal, TEINet:Liu_Luo_Wang_Wang_Tai_Wang_Li_Huang_Lu_2020,STM:Jiang_2019_ICCV, CSN:Tran_2019_ICCV}. These models process the video as a cube to extract spatial-temporal features via the proposed temporal modeling methods. E.g., SlowFast~\citep{SlowFast:feichtenhofer2018slowfast} proposes two pathways with different speed to capture short-range and long-range time dependencies. TSM~\citep{TSM:lin2018temporal} applies a temporal shifting module to exchange information between neighboring frames and TAM~\citep{bLVNetTAM} further enhances TSM by determining the amount of information to be shifted and blended. On the other hand, another thread of work dynamically attempts to select the key frame of an activity for faster recognition~\citep{AdaFrame,LiteEval,ARNet,meng2021adafuse,sun2021dynamic}. E.g., Adaframe~\citep{AdaFrame} employs a policy network to determine whether or not this is a key frame, and the main network only processes the key frames. ARNet~\citep{ARNet} determines what the image resolution should be used to save computations based on the importance of input frame images.
Nonetheless, our approach is fundamentally different from conventional action recognition. It simply uses an image classifier as a video classifier by laying out a video to a super image without explicitly modeling temporal information.

\textbf{Action Recognition with Transformer.} Following the vision transformer (ViT)~\citep{ViT_dosovitskiy2021an}, which demonstrates competitive performance against CNN models on image classification, many recent works attempt to extend the vision transformer for action recognition~\citep{neimark2021videotransformer, VidTr, bertasius2021spacetime, arnab2021vivit, fan2021multiscale}. VTN~\citep{neimark2021videotransformer}, VidTr~\citep{VidTr}, TimeSformer~\citep{bertasius2021spacetime} and ViViT~\citep{arnab2021vivit} share the same concept that inserts a temporal modeling module into the existing ViT to enhance the features from the temporal direction. E.g., VTN~\citep{neimark2021videotransformer} processes each frame independently and then uses a longformer to aggregate the features across frames. On the other hand, divided-space-time modeling in TimeSformer~\citep{TimeSformer_Bertasius_2021vc} inserts a temporal attention module into each transformer encoder for more fine-grained temporal interaction.
MViT~\citep{fan2021multiscale} develops a compact architecture based on the pyramid structure for action recognition. It further proposes a pooling-based attention to mix the tokens before computing the attention map so that the model can focus more on neighboring information.
Nonetheless, our method is straightforward and applies the Swin~\citep{Swin_Liu_2021tq} model to classify super images composed from input frames. 

Note that the joint-space-time attention in TimeSformer~\citep{TimeSformer_Bertasius_2021vc} is a special case of our approach since their method can be considered as flattening all tokens into one plane and then performing self-attention over all tokens. However, the memory complexity of such an approach is prohibitively high, and it is only applicable to the vanilla ViT~\citep{ViT_dosovitskiy2021an} without inductive bias. On the other hand, our \ours is general and applicable to any image classifiers.

\section{Approach}
\label{sec:approach}

\begin{figure}[tb!]
    \centering
    \includegraphics[width=.8\linewidth]{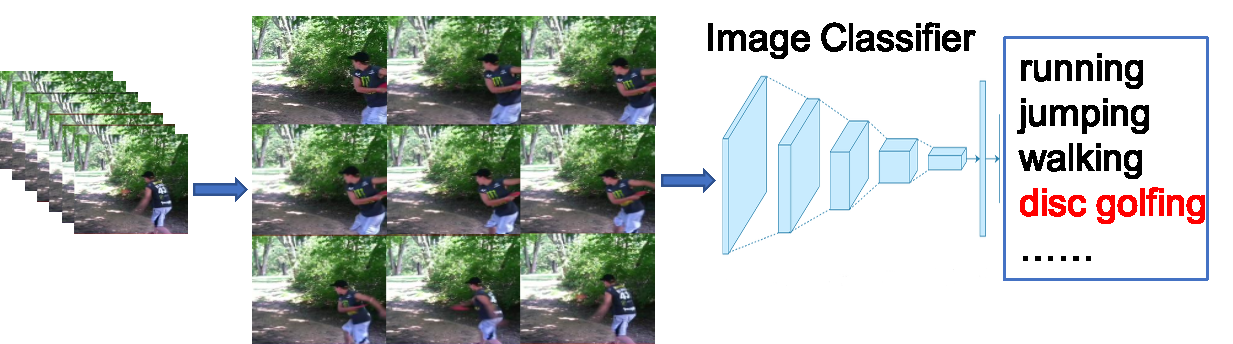}
    \caption{ \textbf{Overview of \ours}. A sequence of input video frames are first rearranged into a super image based on a $3\times3$ spatial layout, which is then fed into an image classifier for recognition. %
    } 
    \label{fig:overview} 
\end{figure}

\subsection{Overview of our Approach}
The key insight of \ours is to turn spatio-temporal patterns in video data into purely 2D spatial patterns in images. Like their 3D counterparts, these 2D patterns may not be visible and recognizable by human. However, we expect they are characteristic of actions and thus identifiable by powerful neural network models.  To that end, we make a sequence of input frame images from a video into a super image, as illustrated in Fig.~\ref{fig:overview}, and then apply an image classifier to predict the label of the video. Note that the action patterns embedded in a super image can be complex and may involve both local (i.e., \textit{spatial information} in a video frame) and global contexts (i.e., \textit{temporal dependencies} across frames). It is thus understandable that effective learning can only be ensured by image classifiers with strong capabilities in modeling short-range and long-range spatial dependencies in super images. For this reason, we explore the recently developed vision transformers based on self-attention to validate our proposed idea. These methods come naturally with the ability to model global image contexts and have demonstrated competitive performance against the best-performed CNN-based approaches on image classification as well as action recognition. Next we briefly describe Swin Transformer~\citep{Swin_Liu_2021tq}, an efficient approach that we choose to implement our main idea in this work.

\textbf{Preliminary.}
The Vision Transformer (ViT) [13] is a purely attention-based classifier borrowed from NLP. It consists of stacked transformer encoders, each of which is featured with a multi-head self-attention module (MSA) and a feed-forward network (FFN). While demonstrating promising results on image classification, ViT uses an isotropic structure and has a quadruple complexity w.r.t image resolution in terms of memory and computation. This significantly limits the application of ViT to many vision applications that requires high-resolution features such as object detection and segmentation.  In light of this issue, several approaches~\citep{Swin_Liu_2021tq, Twins_Chu_2021to, ViL_Zhang_2021tu} have been proposed to perform region-level local self-attention to reduce memory usage and computation, and \Swin is one of such improved vision transformers.

\textbf{Swin Transformer}~\citep{Swin_Liu_2021tq} first adopts a pyramid structure widely used in CNNs to reduce computation and memory. At the earlier layers, the network keeps high image resolution with fewer feature channels to learn fine-grained information. As the network goes deeper, it gradually reduces spatial resolution while expanding feature channels to model coarse-grained information. To further save memory, \Swin limits self-attention to non-overlapping local windows (W-MSA) only.
The communications between W-MSA blocks is achieved through shifting them in the succeeding transformer encoder. The shifted W-MSA is named as SW-MSA. Mathematically, the two consecutive blocks can be expressed as:
\begin{equation} 
\begin{split}
\label{eq:swin}
    \mathbf{y}_k & = \text{W-MSA}(\text{LN}(\mathbf{x}_{k-1})) + \mathbf{x}_{k-1}, \\
    \mathbf{x}_k & = \text{FFN}(\text{LN}(\mathbf{y}_k)) + \mathbf{y}_k, \\
    \mathbf{y}_{k+1} & = \text{SW-MSA}(\text{LN}(\mathbf{x}_{k})) +  \mathbf{x}_{k}, \\
    \mathbf{x}_{k+1} & =\text{FFN}(\text{LN}(\mathbf{y}_{k+1})) +  \mathbf{y}_{k+1},
\end{split}
\end{equation}
where $\mathbf{x}_{l}$ is the features at the $l^{th}$ layer and FFN and LN are feed-forward network and layer normalization, respectively.

\textbf{SIFAR.} In our case, \ours learns action patterns by sliding window, as illustrated in Fig.~\ref{fig:swin}. When the sliding window (blue box) is within a frame, spatial dependencies are learned. On the other hand, when the window (red box) spans across frames, temporal dependencies between them are effectively captured. The spatial pooling further ensures longer-range dependencies across frames captured.

\begin{minipage}[!bt]{\textwidth}
  \begin{minipage}[b]{0.48\textwidth}
    \centering
    \includegraphics[width=.8\linewidth]{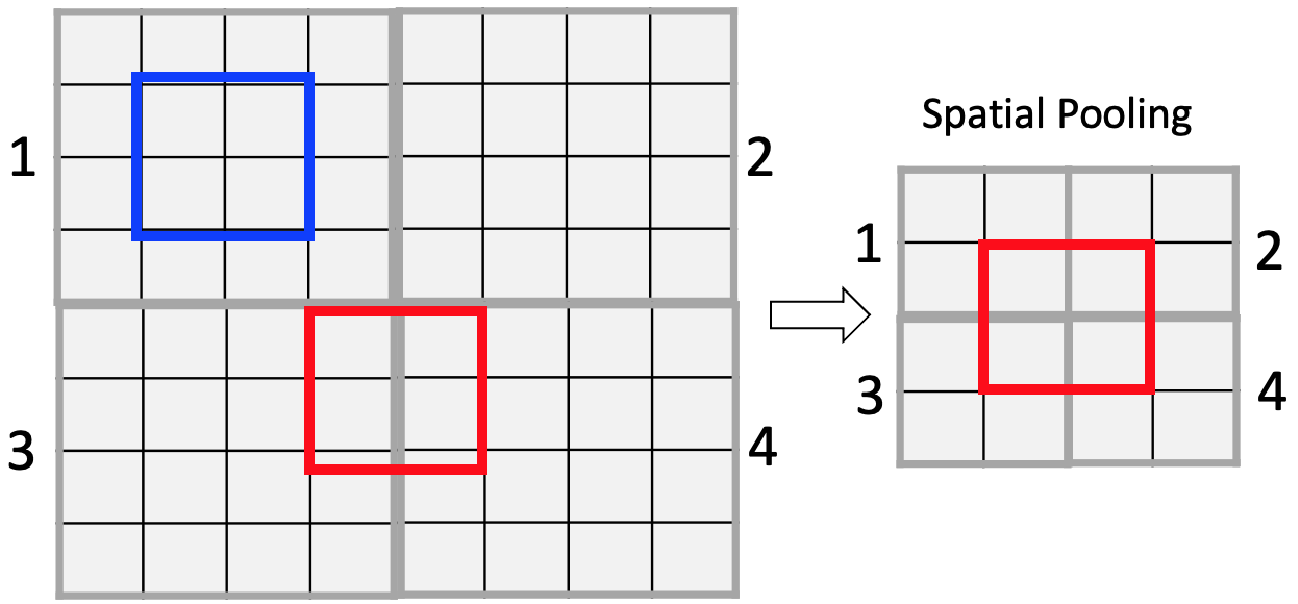}
    \captionof{figure}{\Swin does self-attention in a local window. In SIFAR, when the window (blue box) is within a frame, spatial dependencies are learned within a super image (4 frames here). When the window spans across different frames (red box), temporal dependencies between them are effectively captured. The spatial pooling further ensures longer-range dependencies to be learnt. Best viewed in color.}
    \label{fig:swin} 
  \end{minipage}
  \hfill
  \begin{minipage}[b]{0.48\textwidth}
    \centering
    \includegraphics[width=.75\linewidth]{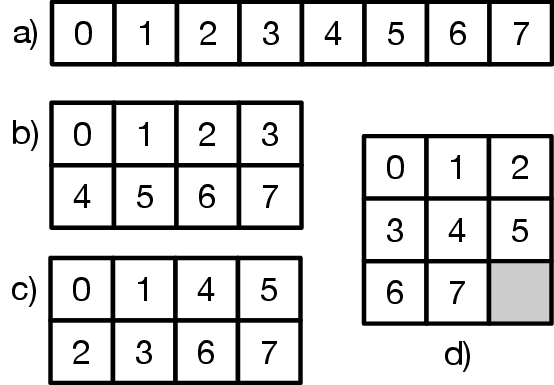}
    \captionof{figure}{\textbf{Grid Layout}. We apply a grid to lay out the input frames. Illustrated here are several possible layouts for 8 frames, i.e., a) $1 \times 8 $, b) and c) $2 \times 4$,  and d) $ 3 \times 3 $, respectively. Empty images are padded at the end if grid is not full. } 
    \label{fig:grid} 
    \end{minipage}
\end{minipage}

\textbf{Creation of Super Image.} Given a set of video frames, we order them by a given layout (Fig.~\ref{fig:grid}) to form a large super image. Different layouts give different spatial patterns for an action class. We hypothesize that a more compact structure such as a square grid may facilitate a model to learn temporal dependencies across frames as such a shape provides the shortest maximum distance between any two images. Given $n$ input frames, we create a super image by placing all the frames in order onto a grid of size $(m-1)\times m$ when $n < (m-1)\times m$ or $m\times m$ when $n\ge (m-1)\times m$ where $m=\lceil \sqrt{n}\rceil$. Empty images are padded at the end if the grid is not full. With this method, for example, 12 frames will be fit into a $3\times4$ grid while 14 frames into a $4\times4$ grid. %
In the default setting, we use a $3 \times 3$ layout for 8 images and a $4 \times 4$ one for 16 images, respectively. There are other spatial arrangements as well (see Fig.~\ref{fig:grid} for more examples). However our experiments empirically show that a square grid performs the best across different datasets.
Note that our approach has linear computational complexity w.r.t the number of input frames. As described above, the size of a super image is $m$ ($m=\lceil \sqrt{n}\rceil$) times as large as the size of a frame image, suggesting that the total number of tokens (or image patches) in Swin grows linearly by $n$. %

\setlength\intextsep{0pt}
\begin{wraptable}{r}{0.5\linewidth}
\centering
    \caption{\textbf{Model architectures of \ours}. The number in a model name indicates the window size used by the model before the last layer. ``B'' means Swin-B. $\dagger$ denotes the models using 16 frames as input and $\ddagger$ indicates the models using a larger input image resolution.}
    \vspace{-1mm}
    \label{table:models}
    \begin{adjustbox}{max width=\linewidth}
    \begin{tabular}{lcccc}
        \toprule
        Model & Frames & Image & FLOPs & Window \\
         & &Size& (G) & Size\\
        \midrule
        \superS & 8 & 224 &138 & \{7,7,7,7\}   \\
        \superM & 8 & 192 &106 &\{12,12,12,\textcolor{blue}{18}\}    \\
        \superL & 8 & 224 &147 & \{14,14,14,\textcolor{blue}{21}\}  \\
        \midrule
        \superMM & 16 & 192 & 189 &\{12,12,12,\textcolor{blue}{24}\} \\
        \superLL & 16 & 224 & 263 &\{14,14,14,\textcolor{blue}{28}\} \\
        \midrule
        \superHH & 8 & 384 & 423 &\{12,12,12,\textcolor{blue}{36}\} \\      
        \bottomrule
    \end{tabular}
    \end{adjustbox}
\end{wraptable}
\textbf{Sliding Window.} As previously mentioned, \Swin performs self-attention within a small local window to save memory. It uses a uniform window size across all layers, and the default window size is 7 in the original paper. Since a larger window leads to more interactions across frames, which is beneficial for \ours to learn long-range temporal dependencies in super images, we slightly modify the architecture of Swin Transformer~\citep{Swin_Liu_2021tq} for it to take different window sizes flexibly in self-attention. In particular, we keep the same window size for all the layers except the last one, whose window is as large as its image resolution, implying a global self-attention including all the tokens. Since the last layer has only two transformer encoders, the computational overhead imposed by an increased window size is quite small, as indicated in Table~\ref{table:models}.

The change of window size may result in adjustment of the input image size as the image resolution at each layer must be divisible by the window size in \Swin. As noted in Table~\ref{table:models}, \superS keeps the vanilla architecture of Swin-B. \superM~is more efficient than \superS~because \superM takes smaller images (192$^{2}$) as input. We demonstrate later in Sec.~\ref{sec:expr} that %
a larger window is critical for \ours to achieve good performance on more temporal datasets such as SSV2.

\mycomment{
\begin{table}[t]
    \centering
    \caption{Model architectures of SuperSWIN}
    \label{table:models}
    \begin{adjustbox}{max width=\linewidth}
    \begin{tabular}{lccccccc}
        \toprule
        Model & Params & FLOPs &#Frames & Image & Window & Dimension  & \#Encoders \\
        &(M) & (G) & &Size & Size & (C) &\\
        \midrule
        \superS & 87.2 & 138& 8 &224 & \{7,7,7,7\} & &   \\
        \superM & 87.3 & 106 & 8& 196 &\{12,12,12,24\} & \{64, 128, 256, 512\} & \{2, 2, 14, 2\}  \\
        \superL & 87.4 & 147 &8&224 &\{14,14,14,28\} &  &  \\
        \midrule
        \superLL & 87.3 & 189 & 16& 320 &\{12,12,12,24\} & \{64, 128, 256, 512\} & \{2, 2, 14, 2\}  \\
        \superLLL & 87.4 & 263 &16&384 &\{14,14,14,28\} &  &  \\             \bottomrule
    \end{tabular}
    \end{adjustbox}
\end{table}

\textbf{Position Embedding.} SWIN applies relative position bias to ... We add an absolute position embedding to each frame.
}

\textbf{Implementation.} Once the spatial layout for the input frames is determined, implementing our idea in PyTorch is as simple as inserting into an image classifier the following line of code, which changes the input of a video to a super image.
{ \small
\begin{Verbatim}[commandchars=\\\{\}]
\textcolor{cyan}{# create a super image with a layout (sh, sw) pre-specified by the user.}
\textcolor{blue}{
x = rearrange(x, 'b c (sh sw) h w -> b c (sh h) (sw w)', sh=sh, sw=sw)
}
\end{Verbatim}
}
The trivial code change described above transforms an image classifier into an video action classifier. Our experiments show that the same training and evaluation protocols for action models can be still applied to the repurposed image classifier.

\section{Experiments}
\label{sec:expr}
\subsection{Datasets and Experimental Setup}
\textbf{Datasets.} We use Kinetics400 (K400)~\citep{Kinetics:kay2017kinetics}, Something-Something V2 (SSV2)~\citep{Something:goyal2017something}, Moments-in-time (MiT)~\citep{Moments:monfort2019moments}, Jester~\citep{Materzynska_2019_ICCV}, and Diving48~\citep{Li_2018_ECCV} datasets in our evaluation. Kinetics400 is a widely-used benchmark for action recognition, which includes $\sim$240k training videos and 20k validation videos in 400 classes.
SSV2 contains 220k videos of 174 types of predefined human-object interactions with everyday objects. This dataset is known for its high temporal dynamics. 
MiT is a fairly large collection of one million 3-second labeled video clips, involving actions not only from humans, but also from animals, objects and natural phenomena. The dataset includes around 800k training videos and 33,900 validation videos in 339 classes. Jester contains actions of predefined hand gestures, with 118,562 and 14,787 training and validation videos over 27 classes, respectively.
Diving48 is an action recognition dataset without representation bias,  which includes 15,943 training videos and 2,096 validation videos over 48 action classes.

\textbf{Training}. We employ \textit{uniform sampling} to generate video input for our models. Such a sampling strategy divides a video into multiple segments of equal length, and  %
has shown to be effective on both Kinetics400 and SSV2~\cite{chen2021deep}.
We train all our models by finetuning a Swin-B model~\citep{Swin_Liu_2021tq} pretrained on ImageNet-21K~\citep{imagenet_deng2009}, except for those SSV2 models, which are fine tuned from the corresponding Kinetics400 models in Table~\ref{table:k400}. 

Our training recipes and augmentations closely follow DeiT~\citep{DeiT_touvron2020}.
First, we apply multi-scale jitter to augment the input~\citep{TSN:wang2016temporal} with different scales and then randomly crop a target input size (e.g. 8$\times$224$\times$224 for \superS).
We then use Mixup~\citep{Mixup_zhang2018} and CutMix~\citep{CutMix_Yun_2019_ICCV} to augment the data further, with their values set to 0.8 and 1.0, respectively. After that, we rearrange the image crops as a super image. Furthermore, we apply drop path~\citep{efficientnet_pmlr_tan_19} with a rate of 0.1, and enable label smoothing~\citep{LabelSmoothing_Szegedy_2016_CVPR} at a rate of 0.1.
All our models were trained using V100 GPUs with 16G or 32G memory. Depending on the size of a model, we use a batch size of 96, 144 or 192 to train the model for 15 epochs on MiT or 30 epochs on other datasets, including 5 warming-up epochs. The optimizer used in our training is AdamW ~\citep{AdamW_loshchilov2018decoupled} with a weight decay of 0.05, and the scheduler is Cosine~\citep{cosine_loshchilov2017} with a base linear learning rate of 0.0001.

\textbf{Inference.} We first scale the shorter side of an image to the model input size %
and then take three crops (top-left, center and bottom-right) for evaluation. The average of the three predictions is used as the final prediction.
We report results by top-1 and top-5 classification accuracy (\%) on validation data, the total computational cost in FLOPs and the model size in number of parameters.

\subsection{Main Results}

\vspace{-2mm}
\setlength\intextsep{0pt}
\begin{wraptable}{r}{0.45\linewidth}
   \centering
   \caption{ \textbf{Comparison with Baseline Methods.} All models use 8 frames as input.}
    \label{table:baseline}
    \vspace{-2mm}
   \begin{adjustbox}{max width=\linewidth}
    \begin{tabular}{l|cccc}
        \toprule
        Model & \multicolumn{2}{c} {SSV2} & \multicolumn{2}{c}{Kinetics400}  \\
         & Top-1 & Top-5 & Top-1 & Top-5\\
         \midrule
         I3D-R50  &61.1&86.5& 72.6 &90.6\\
         TSM-R50  &59.1 &85.6& 74.1 & 91.2\\
         TAM-R50  & 62.0 & 87.3 & 72.2 &90.4\\
         TimeSformer$^*$ &35.9 &71.1& 77.5 &92.5\\
         TimeSformer$^{**}$ &58.7&85.9 & 80.1 &94.4 \\
         \rowcolor{Gray}
        \superS &59.0& 86.0 & 79.6&94.4  \\
        \rowcolor{Gray}
        \superM & 60.8 & 87.3 & 80.0 & 94.5\\
        \rowcolor{Gray}
        \superL &{61.6} & 87.9 &80.2 & 94.4  \\
        \bottomrule
         \multicolumn{5}{l}{\footnotesize $^*$: Swin-B (space only); $^{**}$: Swin-B (divided space-time).} 
    \end{tabular}
    \end{adjustbox}
\end{wraptable}

\mycomment{
\begin{table}[t]
    \centering
    \small
    \begin{tabular}{l|cc|cccc}
        \toprule
        Model &Params & FLOPs & \multicolumn{2}{c} {SSV2} & \multicolumn{2}{c}{Kinetics400}  \\
         &(M) & (G) &Top-1 & Top-5 & Top-1 & Top-5\\
         \midrule
        I3D-R50~\citep{I3D:carreira2017quo} &47.0 & &35.9 &71.1& 72.6 &90.6\\
         TSM-R50~\citep{TSN:wang2016temporal} & & &35.9 &71.1& 77.5 &92.5\\
        TAM-R50~\citep{TSN:wang2016temporal} & & &35.9 &71.1& 77.5 &92.5\\
         TSN-SWIN-B~\citep{TSN:wang2016temporal} & & &35.9 &71.1& 77.5 &92.5\\
         TimeSformer-SWIN-B~\citep{TimeSformer_Bertasius_2021vc}& ? & ? &58.7&85.9 & 80.1 &94.4 \\
         \rowcolor{Gray}
        \superS &87 & 139x3 &56.7& 83.3 & 79.6&94.4  \\
            \rowcolor{Gray}
        \superM & 87& 106x3 & 60.1 & \textbf{86.8} & 80.0 & 94.5\\
     \rowcolor{Gray}
        \superL &87& 147x3 & \textbf{60.6} & 86.7 &\textbf{81.1} & \textbf{94.6}  \\
        \bottomrule
    \end{tabular}
    \caption{\textbf{Comparison with Baseline Methods using SWIN-B as backbone}. All models use 8 frames as input and are finetuned on IN-21K.}
    \label{table:baseline}
\end{table}
}

\textbf{Comparison with Baselines.} We first compare our approach with several representative CNN-based methods including I3D~\citep{I3D:carreira2017quo}, TSM~\citep{TSM:lin2018temporal} and  TAM~\citep{bLVNetTAM}. Also included in the comparison are two TimeSformer models~\citep{TimeSformer_Bertasius_2021vc} based on the same backbone Swin-B~\citep{Swin_Liu_2021tq} as used by our models. All the models considered take 8 frames as input. As can be seen from Table~\ref{table:baseline}, our approach substantially outperforms the CNN baselines on Kinetics400 while achieving comparable results on SSV2. Our approach is also better than TimeSformer on both datasets. These results clearly demonstrate that a powerful image classifier like \Swin can learn expressive spatio-temporal patterns effectively from super images for action recognition. In other words, an image classifier can suffice video understanding without explicit temporal modeling. The results also confirm that a larger sliding window is more helpful in capturing temporal dependencies on temporal datasets like SSV2. Our approach performs global self-attention in the last layer of a model only (see Table~\ref{table:models}). This substantially mitigates the memory issue in training \super models.

\begin{table}[tb]
    \centering
\vspace{-2mm}
    \caption{\textbf{Comparison with Other Approaches on Kinetics400.}}
    \label{table:k400}
    \begin{adjustbox}{max width=0.9\linewidth}
    \begin{tabular}{l|ccccccc}
        \toprule
        Model & \#Frames & Pretrain & Params(M) & FLOPs(G) & Top-1 & Top-5 \\
        \midrule
        TSN-R50~\citep{TSN:wang2016temporal}  & 32 & IN-1K & 24.3 & 170.8$\times$30 & 69.8	& 89.1 \\
        TAM-R50~\citep{bLVNetTAM}  & 32 & IN-1K & 24.4 & 171.5$\times$30 & 76.2	& 92.6 \\
        I3D-R50~\citep{I3D:carreira2017quo}& 32 & IN-1K & 47.0 & 335.3$\times$30 & 76.6 & 92.7   \\
        I3D-R50+NL~\citep{Nonlocal:Wang2018NonLocal} & 32 & IN-1K & $-$ & 282$\times$30 & 76.5 & 92.6   \\
        I3D-R101+NL~\citep{Nonlocal:Wang2018NonLocal} & 32 & IN-1K & $-$ & 359$\times$30 & 77.7 & 93.3  \\
        ip-CSN-152~\citep{CSN:Tran_2019_ICCV} & 32 & $-$ & 32.8 & 109$\times$30 & 77.8 & 92.8  \\
        SlowFast8×8~\citep{SlowFast:feichtenhofer2018slowfast} & 32& $-$ & 27.8 & 65.7$\times$30 & 77.0 & 92.6   \\
        SlowFast8×8+NL~\citep{SlowFast:feichtenhofer2018slowfast} & 32& $-$ & 59.9 & 116$\times$30 & 78.7 & 93.5   \\
        SlowFast16×8+NL~\citep{SlowFast:feichtenhofer2018slowfast} & 64 & $-$ & 59.9 & 234$\times$30 & 79.8 &93.9   \\
        X3D-M~\citep{X3D_Feichtenhofer_2020_CVPR} & 16 & $-$ & 3.8 & 6.2$\times$30 & 76.0 &92.3   \\
        X3D-XL~\citep{X3D_Feichtenhofer_2020_CVPR} & 16 & $-$ & 11.0 & 48.4$\times$30 & 79.1 &93.9   \\
        TPN101~\citep{TPN_yang2020tpn} & 32 & $-$ & & 374$\times$30 & 78.9 & 93.9 \\
        \midrule
        VTN-VIT-B~\citep{neimark2021videotransformer} & 250 & IN-21K & 114.0 &4218$\times$1& 78.6 &93.7 \\  
        VidTr-L~\citep{VidTr} & 32 & IN-21K & $-$ & 351$\times$30 & 79.1 & 93.9 \\
        TimeSformer~\citep{bertasius2021spacetime}& 8 & IN-21K & 121.4 & 196$\times$3 & 78.0 &-   \\
        TimeSformer-HR~\citep{bertasius2021spacetime}& 16 & IN-21K & 121.4 &1703$\times$3 & 79.7 & $-$   \\
        TimeSformer-L~\citep{bertasius2021spacetime}& 96 & IN-21K & 121.4 & 2380$\times$3 & 80.7 & $-$   \\
        ViViT-L~\citep{arnab2021vivit}& 32 & IN-21K &310.8 & 3992$\times$12 & 81.3 &94.7  \\
        MViT-B~\citep{fan2021multiscale}& 16 & $-$ & 36.6 & 70.5$\times$5 & 78.4 &  93.5  \\
        MViT-B~\citep{fan2021multiscale}& 64 & $-$ & 36.6 & 455$\times$9 & 81.2 & 95.1  \\
   \midrule
        \superM& 8 & IN-21K & 87 & 106$\times$3 & 80.0 & 94.5  \\
        \superMM& 16 & IN-21K & 87 & 189$\times$3 & 80.4 & 94.4   \\
        \superL& 8 & IN-21K & 87 & 147$\times$3 & 80.2 & 94.4   \\
        \superLL& 16 & IN-21K & 87 & 263$\times$3 & 81.8&95.2 \\
        \LsuperLL& 16 & IN-21K & 196 & 576$\times$3 & 82.2 & 95.1  \\  
        \superHH& 8 & IN-21K & 87 & 423$\times$3 & 81.6 & 95.2  \\ 
               
        \LsuperHH& 8 & IN-21K & 196 & 944$\times$3 & \textbf{84.2} & \textbf{96.0}  \\         %
        \bottomrule
    \end{tabular} 
    \end{adjustbox}

\end{table}

\textbf{Kinetics400.} Table~\ref{table:k400} shows the results on Kinetics400.
Our 8-frame models (SIFAR-12 and SIFAR-14) achieve $80.0\%$ and $80.2\%$ top-1 accuracies, outperforming all the CNN-based approaches while being more efficient than the majority of them. %
\superLL further gains $\sim1.8\%$ improvement, benefiting from more input frames. Especially, \LsuperHH yields an accuracy of $84.2\%$, the best among all the very recently developed approaches based on vision transformers including TimeSformer~\citep{bertasius2021spacetime} and MViT-B~\citep{fan2021multiscale}. Our proposed approach also offers clear advantages in terms of FLOPs and model parameters compared to other approaches except MViT-B. For example, \superHH has 5$\times$ and 37$\times$ fewer FLOPs than TimeSformer-L and ViViT-L, respectively, while being 1.4$\times$ and 3.6$\times$ smaller in model size.

\textbf{SSV2.}
Table~\ref{table:ssv2} lists the results of our models and the SOTA approaches on SSV2. With the same number of input frames, our approach is $1\sim2\%$ worse than the best-performed CNN methods. However, our approach performs on par with other transformer-based method such as TimeSformer~\citep{TimeSformer_Bertasius_2021vc} and VidTr-L~\citep{VidTr} under the similar setting. Note that ViViT-L~\citep{arnab2021vivit} achieves better results with a larger and stronger backbone ViT-L~\citep{ViT_dosovitskiy2021an}. MViT-B~\citep{fan2021multiscale} is an efficient multi-scale architecture, which can process much longer input sequences to capture fine-grained motion patterns in SSV2 data. Training \super models with more than 16 frames still remains computationally challenging, especially for models like \superL and \LsuperLL, which need a larger sliding window size. Our results suggest that developing more efficient architectures of vision transformer be an area of improvement and future work for \ours to take advantage of more input frames on SSV2. 
 
\begin{table}[tb]
    \centering
\vspace{-2mm}
    \caption{\textbf{Comparison with Other Approaches on SSV2.}}
    \label{table:ssv2}
    \begin{adjustbox}{max width=.8\linewidth}
    \begin{tabular}{l|ccccc}
        \toprule
        Model & \#Frames & Params(M) & FLOPs(G) & Top-1 & Top-5 \\
        \midrule
        TAM-R50~\citep{bLVNetTAM} &8 & 24.4 & 42.9$\times$6 & 62.8 & 87.4 \\
        TAM-R50~\citep{bLVNetTAM} &32& 24.4 & 171.5$\times$6 & 63.8	& 88.3 \\
        I3D-R50~\citep{I3D:carreira2017quo}&8& 47.0 & 83.8$\times$6 &  61.1 & 86.5 \\
        I3D-R50~\citep{I3D:carreira2017quo}&32 & 47.0 & 335.3$\times$6 & 62.8 & 88.0   \\
        TSM-R50~\citep{TSM:lin2018temporal}& 8 & 24.3 & 32$\times$6 & 59.1 & 85.6 \\
     TSM-R50~\citep{TSM:lin2018temporal}&16 & 24.3 & 65$\times$6 &63.4 & 88.5 \\
     TPN-R50~\citep{TPN_yang2020tpn} & 8 & $-$ & $-$ & 62.0 & $-$ \\
        TAM-bLR101~\citep{bLVNetTAM} &64 & 40.2 & 96.4$\times$1 & 65.2 & 90.3   \\
        MSNet~\citep{MotionSqueeze_kwon2020motionsqueeze}& 16& 24.6 & 67$\times$1 & 64.7 & 89.4  \\
        STM~\citep{jiang2019stm}& 16& 24.0 & 67$\times$30 & 64.2 & 89.8  \\
        TEA~\citep{TEINet:Liu_Luo_Wang_Wang_Tai_Wang_Li_Huang_Lu_2020}  &16 & $-$ & 70$\times$30 & 65.1 & 89.9  \\
        \midrule
        TimeSformer~\citep{bertasius2021spacetime}&8&121.4 & 196$\times$3 & 59.5 & $-$   \\
        TimeSformer-HR~\citep{bertasius2021spacetime}&16 &121.4 & 1703$\times$3 & 62.5 & $-$   \\
        ViViT-L~\citep{arnab2021vivit}& 32 & 100.7 & $-$ & 65.4 & 89.8  \\
        VidTr-L~\citep{VidTr} & 32 & $-$ & $-$ & 60.2 & $-$ \\
        
        MViT-B~\citep{fan2021multiscale}& 16 & 36.6 & 70.5$\times$3 & 64.7 & 89.2   \\
        MViT-B~\citep{fan2021multiscale}& 64 & 36.6 & 455$\times$3 & \textbf{67.7} & \textbf{90.9}   \\
   \midrule
        \superM &8 & 87 & 106$\times$3 & 60.8 & 87.3   \\
        \superMM &16 & 87 & 189$\times$3 & 61.4 & 87.4   \\       
        \superL&8 & 87 & 147$\times$3 & 61.6 & 87.9   \\
        \superLL&16 & 87 & 263$\times$3 & 62.6& 88.5  \\
        \LsuperLL  & 16 & 196 & 576$\times$3 & 64.2 & 88.4  \\         
        \bottomrule
    \end{tabular} 
    \end{adjustbox}
\vspace{-5mm}
\end{table}

\begin{table}[tb]
    \centering
    \caption{\textbf{Comparison with Other Methods on MiT.}}
    \vspace{-3mm}
    \label{table:mit}
    \begin{adjustbox}{max width=0.8\linewidth}
    \begin{tabular}{l|cc}
        \toprule
        Model & Top-1 & Top-5 \\
        \midrule
        TRN-Incpetion~\citep{TRN:zhou2018temporal} & 28.3  & 53.9 \\
        TAM-R50~\citep{bLVNetTAM} & 30.8 & 58.2   \\
        I3D-R50~\citep{chen2021deep}& 31.2 & 58.9  \\
        SlowFast-R50-8$\times$8~\citep{SlowFast:feichtenhofer2018slowfast} & 31.2 & 58.7   \\
        CoST-R101~\citep{CollaborativeST:Li_2019_CVPR} & 32.4 & 60.0 \\
        SRTG-R3D-101~\citep{SRTG} & 33.6 & 58.5  \\
        AssembleNet~\citep{DBLP:journals/corr/abs-1905-13209} &  33.9 & 60.9 \\
        ViViT-L~\citep{arnab2021vivit} & 38.0 & 64.9  \\
        \superHH & 39.9 & 69.2 \\ %
        \LsuperHH & \textbf{41.9} & \textbf{70.3} \\ %
        \bottomrule
    \end{tabular}
    \end{adjustbox}
    \vspace{-2mm}
\end{table}

\textbf{MiT.}
MiT is a large diverse dataset containing label noise. As seen from Table~\ref{table:mit}, with the same backbone ViT-L, \LsuperHH is $\sim$4\% better than ViViT-L~\citep{arnab2021vivit}, and outperforms AssembelNet~\citep{AssembleNet:Ryoo2020AssembleNet_ICLR2020} based on neural architecture search by a considerable margin of 8\%. 

\textbf{Jester and Diving48.} We further evaluate our proposed approach on two other popular benchmarks: Jester~\citep{Materzynska_2019_ICCV} and Diving48~\citep{Li_2018_ECCV}. Here we only consider the best single models from other approaches for fair comparison. As shown in Table~\ref{table:jester_diving}, \ours achieves competitive results again on both datasets, surpassing all other models in comparison. %
Note that the Diving48 benchmark contains videos with similar background and objects but different action categories, and is generally considered as an unbiased benchmark. Our model \superLL outperforms TimeSformer-L by a large margin of 6\% on this challenging Diving48 dataset.

\mycomment{
\begin{table}[tb]
    \centering
    \caption{\textbf{Comparison with Other Approaches on MiT and Jester.}}
    \label{table:mit_jester}
    \begin{subtable}[h]{0.48\textwidth}
        \begin{table}[tb]
    \centering
    \caption{\textbf{Comparison with Other Methods on MiT.}}
    \vspace{-3mm}
    \label{table:mit}
    \begin{adjustbox}{max width=0.8\linewidth}
    \begin{tabular}{l|cc}
        \toprule
        Model & Top-1 & Top-5 \\
        \midrule
        TRN-Incpetion~\citep{TRN:zhou2018temporal} & 28.3  & 53.9 \\
        TAM-R50~\citep{bLVNetTAM} & 30.8 & 58.2   \\
        I3D-R50~\citep{chen2021deep}& 31.2 & 58.9  \\
        SlowFast-R50-8$\times$8~\citep{SlowFast:feichtenhofer2018slowfast} & 31.2 & 58.7   \\
        CoST-R101~\citep{CollaborativeST:Li_2019_CVPR} & 32.4 & 60.0 \\
        SRTG-R3D-101~\citep{SRTG} & 33.6 & 58.5  \\
        AssembleNet~\citep{DBLP:journals/corr/abs-1905-13209} &  33.9 & 60.9 \\
        ViViT-L~\citep{arnab2021vivit} & 38.0 & 64.9  \\
        \superHH & 39.9 & 69.2 \\ %
        \LsuperHH & \textbf{41.9} & \textbf{70.3} \\ %
        \bottomrule
    \end{tabular}
    \end{adjustbox}
    \vspace{-2mm}
\end{table}

    \end{subtable}
    \quad
    \begin{subtable}[h]{0.48\textwidth}
            \centering
    \caption{{Jester}}
    \label{table:jester}
    \begin{adjustbox}{max width=\linewidth}
    \begin{tabular}{l|cc}
        \toprule
        Model & Top-1 & Top-5  \\
         \midrule
        TSN-Inception~\citep{TSN:wang2016temporal}&  95.0 & 99.9    \\
        TRN-Inception~\citep{TRN:zhou2018temporal} & 95.3 & $-$ \\
        TSM-R50~\citep{TSM:lin2018temporal} & 95.0 & 99.9    \\
        PAN-R50~\citep{PAN} & 99.6 & 99.8\\
        STM-R50\citep{STM:Jiang_2019_ICCV} & 96.7 & 99.9 \\
        I3D-R50~\citep{I3D:carreira2017quo} & 96.4 & $-$ \\
        TAM-R50~\citep{bLVNetTAM} & 96.4 & $-$ \\
        SlowFast-R50-8$\times$8 \citep{SlowFast:feichtenhofer2018slowfast} & 96.8 & $-$ \\
        \superMM & {97.2} & {99.9}    \\
        \superLL & \textbf{97.2} & \textbf{99.9}  \\
        \bottomrule
    \end{tabular}
    \end{adjustbox}

    \end{subtable}
    \end{table}
}

\begin{table}[tb]
    \centering
    \caption{\textbf{Comparison with Other Approaches on Jester and Diving48.}}
    \label{table:jester_diving}
    \begin{subtable}[h]{0.48\textwidth}
            \centering
    \caption{{Jester}}
    \label{table:jester}
    \begin{adjustbox}{max width=\linewidth}
    \begin{tabular}{l|cc}
        \toprule
        Model & Top-1 & Top-5  \\
         \midrule
        TSN-Inception~\citep{TSN:wang2016temporal}&  95.0 & 99.9    \\
        TRN-Inception~\citep{TRN:zhou2018temporal} & 95.3 & $-$ \\
        TSM-R50~\citep{TSM:lin2018temporal} & 95.0 & 99.9    \\
        PAN-R50~\citep{PAN} & 99.6 & 99.8\\
        STM-R50\citep{STM:Jiang_2019_ICCV} & 96.7 & 99.9 \\
        I3D-R50~\citep{I3D:carreira2017quo} & 96.4 & $-$ \\
        TAM-R50~\citep{bLVNetTAM} & 96.4 & $-$ \\
        SlowFast-R50-8$\times$8 \citep{SlowFast:feichtenhofer2018slowfast} & 96.8 & $-$ \\
        \superMM & {97.2} & {99.9}    \\
        \superLL & \textbf{97.2} & \textbf{99.9}  \\
        \bottomrule
    \end{tabular}
    \end{adjustbox}

    \end{subtable}
    \quad
    \begin{subtable}[h]{0.48\textwidth}
            \centering
    \caption{{Diving48}}
    \label{table:diving48}
  \begin{adjustbox}{max width=\linewidth}
    \begin{tabular}{l|cc}
        \toprule
        Model & Top-1 & Top-5  \\
         \midrule
        TimeSformer~\citep{bertasius2021spacetime}&  74.9 & $-$    \\
         TimeSformer-HR~\citep{TimeSformer_Bertasius_2021vc}& 78.0 & $-$ \\
        TimeSformer-L~\citep{TimeSformer_Bertasius_2021vc} & 81.0 & $-$    \\
        SlowFast~\citep{SlowFast:feichtenhofer2018slowfast} & 77.6 & $-$    \\
        \superMM & {85.3} & {98.3}    \\
        \superLL & \textbf{87.3} & \textbf{98.8}  \\
        \bottomrule
    \end{tabular}
    \end{adjustbox}

    \end{subtable}
    \vspace{-5mm}
    \end{table}

\textbf{Classification by CNNs.}
We also test our proposed approach using the ResNet image classifiers on both SSV2 and Kinetics400 datasets. For fairness, the ResNet models are pretrained on ImageNet-21K. Table~\ref{table:cnn} shows the results. Our models clearly outperform the traditional CNN-based models for action recognition on Kinetics400. Especially, with a strong backbone R152x2 (a model 2$\times$ wider than Resnet152), SIFAR-R152x2 achieves a superior accuracy of 79.0\%, which is surprisingly comparable to the best CNN results (SlowFast16×8+NL: 79.8\%) reported in Table~\ref{table:k400}.

\setlength\intextsep{0pt}
\begin{wraptable} {l}{0.5\textwidth}
\mycomment{
\begin{table}[t]
   \centering
   \caption{\textbf{\ours based on CNN}. The base denotes that Mixup, CutMix and drop-path are disabled; the Aug. follows our training setting. All results are evaluated under three crops.}
    \label{table:cnn}
    \begin{tabular}{l|ccccc}
        \toprule
        Model & \# Frames & \multicolumn{2}{c} {SSV2} & \multicolumn{2}{c}{Kinetics400}  \\
         & & Base & Aug & Base & Aug \\
         \midrule
         I3D-R50~\citep{I3D:carreira2017quo}   & 8 &61.1& - & 72.6 & - \\
         TSM-R50~\citep{TSN:wang2016temporal} & 8 &59.1 &-& 74.1 &-\\
         TAM-R50~\citep{bLVNetTAM} & 8 & 62.0 & - & 72.2 &-\\
        \super-R50 & 8 & 48.6  & 50.8  & 73.5  & 73.2  \\
        \super-R50 & 16 & 46.8 & 49.0  & 73.3 & 73.5  \\
        \super-R101 & 8 & 53.7 & 56.3  & 75.5 & 76.6  \\
        \super-R101 & 16 & 52.5 & 56.2 & 75.8 & 77.4   \\
        \super-R152$\times$2$^*$ & 8 & 53.5 & 58.2  & 76.0 & 79.0  \\
        \super-R152$\times$2$^*$ & 16 & 54.5 & 58.6 & 77.8 & 80.0   \\
        \rowcolor{Gray}
        \superS  & 8 & 53.4 & 56.7 &  78.7 & 79.6                \\
        \bottomrule
         \multicolumn{6}{l}{\footnotesize $^*$: a model two times wider than R152} 
        
    \end{tabular}
\end{table}
}

   \centering
   \caption{CNN-based \super Results}
    \label{table:cnn}
   \begin{adjustbox}{max width=\linewidth}
    \begin{tabular}{l|ccc}
        \toprule
        Model & \# Frames & SSV2 & {Kinetics400}  \\
         \midrule
         I3D-R50~\citep{I3D:carreira2017quo}   & 8 &61.1& 72.6 \\
         TSM-R50~\citep{TSN:wang2016temporal} & 8 &59.1 & 74.1 \\
         TAM-R50~\citep{bLVNetTAM} & 8 & 62.0 & 72.2 \\
                 \midrule
        \super-R50 & 8   & 50.8  & 73.2  \\
        \super-R101 & 8  & 56.3  & 76.6  \\
        \super-R152$\times$2$^*$ & 8 & 58.2   & 79.0  \\
        \midrule
        \super-R50-C7 & 8 & 54.4 (\textcolor{red}{+3.6}) & 74.4 (\textcolor{red}{+1.2})  \\\super-R50-C11 & 8 & 55.2 (\textcolor{red}{+4.2}) & 74.5 (\textcolor{red}{+1.3})  \\
        \super-R50-C21 & 8 & 55.8 (\textcolor{red}{+5.0}) & 74.8 (\textcolor{red}{+1.6})  \\
        \super-R50-C21-11 & 8 & 57.6 (\textcolor{red}{+6.8}) & 75.1 (\textcolor{red}{+1.9})  \\
        \super-R101-C21 & 8 &58.1 (\textcolor{red}{+1.8})   & 77.7 (\textcolor{red}{+1.1})  \\
        \super-R101-C21-11 & 8 & 59.6 (\textcolor{red}{+3.3}) & 77.5 (\textcolor{red}{+0.9})    \\

        \bottomrule
         \multicolumn{4}{l}{\footnotesize $^*$: a model two times wider than R152} 
        
    \end{tabular}
    \end{adjustbox}

\mycomment{
\begin{table}[t]
   \centering
   \caption{\textbf{\ours based on CNN}. The base denotes that Mixup, CutMix and drop-path are disabled; the Aug. follows our training setting. All results are evaluated under three crops.}
    \label{table:cnn}
    \begin{tabular}{l|ccc}
        \toprule
        Model & \# Frames & \multicolumn{2}{c}{Kinetics400}  \\
         & & Base & Aug \\
         \midrule
         I3D-R50~\citep{I3D:carreira2017quo}   & 8 & 72.6 & - \\
         TSM-R50~\citep{TSN:wang2016temporal} & 8 & 74.1 &-\\
         TAM-R50~\citep{bLVNetTAM} & 8  & 72.2 &-\\
        \super-R50 & 8 & 73.5  & 73.2  \\
        \super-R50 & 16  & 73.3 & 73.5  \\
        \super-R101 & 8  & 75.5 & 76.6  \\
        \super-R101 & 16 & 75.8 & 77.4   \\
        \super-R152$\times$2$^*$ & 8 & 76.0 & 79.0  \\
        \super-R152$\times$2$^*$ & 16 & 77.8 & 80.0   \\
        \bottomrule
         \multicolumn{6}{l}{\footnotesize $^*$: a model two times wider than R152} 
        
    \end{tabular}
\end{table}
}
\end{wraptable}
On SSV2, the results of CNN-based \super are less satisfactory but reasonable. This is because 3x3 convolutions are local with a small receptive field, thus failing to capturing long-range temporal dependencies in super images. We hypothesize that a larger kernel size with a wider receptive field may address this limitation and potentially improve the performance of CNN-based SIFAR models. To validate this, we perform additional experiments by adding one or two more residual blocks to the end of ResNet models with larger kernel sizes, i.e. replacing the second convolution in those new blocks by a 7x7, 11x11 or 21x21 kernel. These models are indicated by names ending with  ``C7'' (7x7), ``C11'' (11x11) or ``C21'' (21x21) in Table~\ref{table:cnn}. As seen from the table, using larger kernel sizes consistently improves the results on both ResNet50 and ResNet101 models. For example, we obtain an absolute 5.0\% improvement over original ResNet50 and 2.0\% over original ResNet101 respectively, using one more block with a kernel size of 21x21. When adding another block with a kernel size of 11x11 (i.e. SIFA-R50-C21-11 and SIFA-R101-C21-11), it further boosts the performance up to 6.8\% with ResNet50 and 2.7\% with ResNet101. These results strongly suggest that expanding the receptive field of CNNs be a promising direction to design better CNN-based SIFAR models. 

\subsection {Ablation Studies}
\vspace{-2mm}

\textbf{How does an image layout affect the performance?} The layout of a super image determines how spatio-temporal patterns are embedded in it. 
To analyze this, we trained a SIFAR model on SSV2 for each layout illustrated in Fig.~\ref{fig:grid}. As shown in Table~\ref{table:layout}, a strip layout performs the worst while a grid layout produces the best results, which confirms our hypothesis.

\textbf{Does absolute positioning embedding help?}
Swin paper~\citep{Swin_Liu_2021tq} shows that when relative position bias are added, Absolute Position Embedding (APE) is only moderately beneficial for classification, but not for object detection and segmentation. They thus conclude that inductive bias that encourages certain translation invariance is still important for vision tasks. To find out whether or not APE is effective in our proposed approach, we add APE to each frame rather than each token. The results in Table~\ref{table:ape} indicate that APE slightly improves model accuracy on SSV2, but is harmful to Kinetics400. In our main results, we thus apply APE to SSV2 only.

\begin{table}[tb]
    \centering
    \caption{{\textbf{Ablation Study.} The effects of each component on model accuracy.}}
    \vspace{-1mm}
    \label{table:abl_all}
    \begin{subtable}[h]{0.48\textwidth}
        \centering
        \caption{{Super Image Layout.}
        (\superM on SSV2)}
        \vspace{-2mm}
        \label{table:layout}
        \begin{adjustbox}{max width=\linewidth}
        \begin{tabular}{l|cc}
            \toprule
            Layout & Top-1 & Top-5 \\
            \midrule
            1$\times$8~(Fig.~\ref{fig:grid}a) &44.4 &74.4 \\
            2$\times$4~(Fig.~\ref{fig:grid}b) &58.6 &85.5  \\
            2$\times$4~(Fig.~\ref{fig:grid}c) &58.1 &85.1  \\
            3$\times$3~(Fig.~\ref{fig:grid}d) &60.8 & 87.3 \\
            \bottomrule
       \end{tabular}
       \end{adjustbox}
    \end{subtable}
    \quad
    \begin{subtable}[h]{0.48\textwidth}
        \centering
        \caption{{{Absolute Positioning Embedding}.}}
        \vspace{-2mm}
        \label{table:ape}
        \begin{adjustbox}{max width=\linewidth}
        \begin{tabular}{l|cccc}
            \toprule
            Model &\multicolumn{2}{c}{SSV2} & \multicolumn{2}{c}{Kinetics400}  \\
             & w/ APE & w/o APE & w/ APE&w/o APE \\
             \midrule
            \superS &56.6 &56.4 & 79.7 & 79.6\\
            \superM &60.8 &59.5 & 79.7 &80.0\\
            \superL &61.6 &60.1 &80.0  &80.2\\
            \bottomrule
       \end{tabular}
       \end{adjustbox}
    \end{subtable}
    \vspace{-4mm}
\end{table}

\setlength\intextsep{0pt}
\begin{wraptable} {r}{0.35\textwidth}
    \centering
    \caption{\small \textbf{Effects of temporal order.}}
    \vspace{-3mm}
    \label{table:temporal-order}
    \begin{tabular}{l|cc}
        \toprule
        Order & Kinetics400 & SSV2  \\
         \midrule
        normal&  80.0 & 60.8    \\
        reverse& 79.8 & 23.9 \\
        random&  79.7& 39.4   \\
    \bottomrule
    \end{tabular}
\end{wraptable}

\textbf{Does the temporal order of input matter?} %
We evaluate  the \superM model (trained with input frames of normal order) by using three types of input with different temporal orders, i.e the reverse, random and normal.
As shown in Table~\ref{table:temporal-order}, \super is not sensitive to the input order on Kinetics400, whereas on SSV2, changing the input order results in a significant performance drop. This is consistent with the finding in the S3D paper~\citep{S3D}, 
indicating that for datasets like SSV2 where there are visually similar action categories, the order of input frames matters in model learning.

\begin{figure}[tbh!]
    \centering
    \vspace{2mm}
    \includegraphics[width=0.90\linewidth]{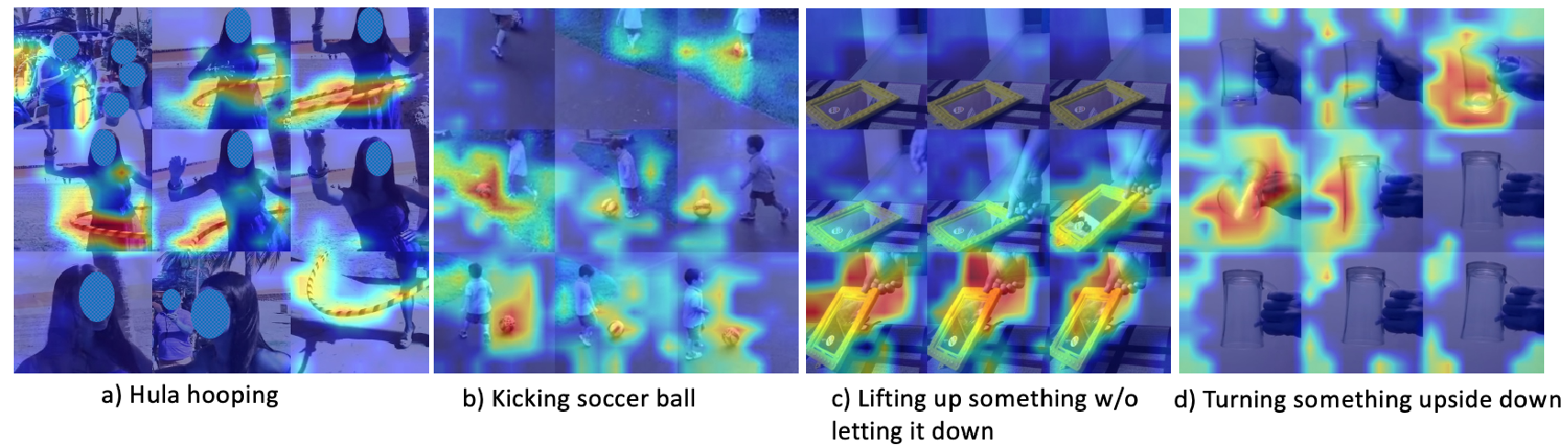} \vspace{-1mm}
    \caption{\textbf{Visualization by Ablation CAM~\citep{desai2020ablationcam}}}
    \label{fig:vis} 
  \vspace{2mm}
\end{figure}

\textbf{What does \ours learn?}
We apply ablation CAM~\citep{desai2020ablationcam}, an image model interpretability technique, to understand what our models learn. Fig.~\ref{fig:vis} shows the Class Activation Maps (CAM) of $4$ actions correctly predicted by \superM. Not surprisingly, the model learns to attend to objects relevant to the target action such as the hula hoop in a) and soccer ball in b). In c) and d), the model seems to correctly focus on where meaningful motion happens.

\vspace{-1mm}
\section{Conclusion}
\label{sec:conclusion}
\vspace{-1mm}

In this paper, we have presented a new perspective for action recognition by casting the problem as an image recognition task. Our idea is simple but effective, and with one line of code to transform an sequence of input frames into a super image, it can re-purpose any image classifier for action recognition. We have implemented our idea with both CNN-based and transformer-based image classifiers, both of which show promising and competitive results on several popular publicly available video benchmarks. Our extensive experiments and results show that applying super images for video understanding is an interesting direction worth further exploration.

\mycomment{
\textbf{Boarder Impact.} 
Our work introduces a new perspective for action recognition, which simply applies an image classifier for recognizing activity. By doing so, this work has an important positive impact that it enables the reusability of research on image classification, like the ablation CAM we showed in the paper instead of inventing new technique to explain an action recognition model. Negative impacts of this work are not easy to predict, however, it shares many of the pitfalls
associated with deep classification models. E.g., the proposed video classification systems have negative implications on privacy and could be used by malicious actors or governments to infringe on the privacy of citizens. Future research into private and ethical aspects of visual recognition is an important direction.
}

\bibliography{reference,more_reference}
\bibliographystyle{iclr2022_conference}

\end{document}